\documentclass[11pt]{article}

\usepackage[margin=1in]{geometry}
\usepackage{times}
\usepackage{amsmath, amssymb}
\usepackage{graphicx}
\usepackage{booktabs}
\usepackage{hyperref}
\usepackage{microtype}
\usepackage{float}
\usepackage{graphicx}
\graphicspath{{./}{figures/}}
\hypersetup{
    colorlinks=true,
    linkcolor=black,
    citecolor=black,
    urlcolor=blue
}

\title{
From Scarcity to Scale: A Release-Level Analysis of the Pashto Common Voice Dataset
}

\author{
Jandad Jahani \\
O.P. Jindal Global University \\
\texttt{jandadjahani@gmail.com}
\\[0.8em]
Mursal Dawodi \\
Technical University of Munich  \\
\texttt{mursal.dawodi@tum.de}
\\[0.8em]
Jawid Baktash \\
Technical University of Munich  \\
\texttt{jawid.baktash@tum.de}
}

\date{}  % No date

\begin{document}

\maketitle

\begin{abstract}
Large, openly licensed speech datasets are essential for building automatic speech recognition (ASR) systems, yet many widely spoken languages remain underrepresented in public resources. Pashto, spoken by more than 60 million people, has historically lacked large-scale openly licensed speech data suitable for modern ASR development.

This paper presents a release-level analysis of the Pashto component of the Mozilla Common Voice corpus, focusing on version 24.0 (December 2025) and contextualizing trends across major releases. We document rapid growth from 1.49 recorded hours in mid-2023 to 2,768.7 total hours in 2025, including 975.89 validated hours available for supervised ASR training.

Beyond scale, we analyze validation throughput, contributor participation inequality, demographic metadata completeness, and sentence-level concentration in the validated subset. We find that participation is extremely concentrated (Gini = 0.941), age representation is strongly skewed toward young adults, and 41.97\% of clips lack self-reported gender labels, limiting subgroup auditing based on metadata. At the textual level, prompt reuse is moderate: 35.88\% of unique sentences account for 50\% of validated clips, suggesting that structural concentration is driven primarily by uneven contributor activity rather than dominance of a small prompt set.

These results provide a quantitative audit of a rapidly scaling low-resource speech corpus and highlight practical priorities for improving dataset maturity, including expanded validation capacity and broader demographic participation.
\end{abstract}

\section*{Keywords}
Pashto; Automatic Speech Recognition (ASR); Low-Resource Languages; Crowdsourced Speech Data; Common Voice

\section{Introduction}

Large-scale speech datasets are central to contemporary automatic speech recognition (ASR) systems. Over the past decade, major improvements in ASR performance have been driven by the availability of openly licensed corpora that support large-scale training and reproducible benchmarking. However, this progress has been uneven across languages. Many widely spoken languages remain underrepresented in publicly accessible speech resources, contributing to disparities in research attention and technological performance~\cite{ref1,ref2}.

Pashto, an Indo-Iranian language spoken by more than 60 million people primarily in Afghanistan and Pakistan, illustrates this imbalance. Despite its large speaker population, Pashto has historically been considered low-resource in speech technology research due to the limited availability of openly licensed, large-scale speech corpora suitable for modern ASR development~\cite{ref1}. As a result, Pashto speech technologies have lagged behind those developed for higher-resource languages.

Crowdsourced data collection has emerged as a scalable solution to speech resource scarcity. The Mozilla Common Voice initiative is one of the largest implementations of this model, enabling volunteers to record and validate speech data under a CC0 1.0 public-domain dedication~\cite{ref3}. By reducing legal and financial barriers, Common Voice has expanded multilingual speech resources substantially, including for languages that previously lacked publicly accessible corpora.

While dataset scale is often treated as a primary indicator of progress, size alone does not determine representational adequacy. Prior research has shown that ASR systems are sensitive to demographic composition and recording conditions, and that imbalances in training data can lead to systematic performance disparities across user groups~\cite{ref5,ref6}. Consequently, evaluating a speech corpus requires examining not only its total duration, but also its participation structure, validation dynamics, and metadata completeness.

This paper presents a release-level analysis of the Pashto component of the Mozilla Common Voice corpus, focusing primarily on version 24.0 (December 2025). Within three years, the dataset expanded from fewer than two hours of recorded speech to 2,768.7 total hours, including 975.89 validated hours suitable for ASR training. The v24.0 release comprises 2,407,799 clips contributed by 6,654 speakers and represents, to our knowledge, the largest openly licensed Pashto speech dataset currently available.

Rather than evaluating this growth solely in terms of scale, we analyze how expansion interacts with validation throughput, contributor concentration, demographic metadata, and sentence-level distribution. We show that substantial increases in total duration coexist with pronounced participation inequality, incomplete demographic reporting, and structured concentration patterns at the speaker level. These characteristics have implications for downstream ASR robustness and fairness.

Specifically, this paper makes the following contributions:

\begin{enumerate}
    \item \textbf{Longitudinal Dataset Analysis.} We document the evolution of the Pashto Common Voice corpus across major releases and characterize its transition from scarcity to large-scale availability.
    
    \item \textbf{Structural Composition Assessment.} We quantify validation status, contributor participation inequality, demographic metadata distribution, and sentence-level concentration within the validated subset.
    
    \item \textbf{Implications for ASR Development.} We discuss how participation inequality and metadata incompleteness may influence model robustness and demographic generalization.
\end{enumerate}

By providing a transparent quantitative audit of a rapidly scaling low-resource speech corpus, this study contributes to more informed and responsible development of Pashto ASR systems.

\section{Related Work}

Advances in automatic speech recognition (ASR) have historically depended on the availability of large-scale, well-documented training corpora~\cite{ref1}. Early benchmark datasets such as LibriSpeech provided openly licensed, reproducible resources that accelerated neural ASR research~\cite{ref9}. However, these corpora were linguistically narrow, primarily based on read English audiobooks, and reflected limited demographic and acoustic diversity.

Subsequent multilingual efforts expanded both scale and language coverage. Multilingual LibriSpeech (MLS) and VoxPopuli leveraged audiobooks, parliamentary proceedings, and broadcast recordings to construct aligned multilingual corpora spanning dozens of languages~\cite{ref8,ref10}. These datasets have advanced multilingual benchmarking and cross-lingual transfer research. At the same time, their reliance on institutional or semi-formal speech sources constrains variation in speaking style, recording conditions, and informal usage.

For languages historically classified as low-resource, data scarcity remains a central barrier to ASR development~\cite{ref1,ref2}. Beyond technical constraints, scholars have also emphasized structural inequalities in speech and language technology ecosystems, including data ownership, access, and representational imbalance~\cite{ref4}. Crowdsourced collection has therefore emerged as a scalable alternative to institutional recording pipelines.
 The Mozilla Common Voice initiative is among the largest implementations of this model, enabling volunteers to record and validate speech data under a CC0 1.0 public-domain dedication~\cite{ref3,ref12}. Prior studies primarily evaluate Common Voice as a training resource and benchmark corpus~\cite{ref3}, focusing on model performance rather than detailed structural analysis of individual language components.

Parallel work has documented demographic disparities in ASR systems. Empirical studies demonstrate performance differences across gender, dialect, and accent, often linked to imbalances in training data composition~\cite{ref5,ref6}. These findings underscore the importance of examining dataset structure, demographic metadata availability, and participation patterns when evaluating speech resources.

Despite growing multilingual coverage, systematic release-level analyses of individual language subsets within large crowdsourced speech corpora remain limited, particularly for low-resource languages. For Pashto specifically, publicly documented speech resources have historically been small, restricted by licensing, or insufficiently described for large-scale reproducible research. While Pashto has appeared in multilingual ASR experiments, no prior work has provided a longitudinal structural analysis of the Pashto component of the Common Voice corpus.

This study addresses this gap by shifting attention from model performance to dataset composition. We conduct a release-level analysis of the Pashto subset, quantifying participation structure, validation dynamics, demographic metadata distribution, and sentence-level concentration patterns within a rapidly scaling corpus.

\noindent
To contextualize the scale of the Pashto v24.0 release within the broader Common Voice ecosystem, 
Table~\ref{tab:cv_cross_language_v24} presents a cross-language comparison using release-level statistics from the Common Voice dataset cards. The comparison includes total hours, validated hours, validation rate, 
number of clips, and number of contributors.

\begin{table}[H]
\centering
\small
\setlength{\tabcolsep}{6pt}
\begin{tabular}{lrrrrr}
\hline
\textbf{Language (v24.0)} & \textbf{Total Hours} & \textbf{Validated Hours} & \textbf{Validation Rate} & \textbf{Clips} & \textbf{Speakers} \\
\hline
Pashto (ps)   & 2768.70 & 975.89 & 35.2\% & 2{,}407{,}799 & 6{,}654 \\
Persian (fa)  & 428.61  & 373.66 & 87.2\% & 390{,}134   & 4{,}639 \\
Urdu (ur)     & 302.02  & 81.48  & 27.0\% & 252{,}899   & 498 \\
Uzbek (uz)    & 265.45  & 100.69 & 37.9\% & 229{,}837   & 2{,}281 \\
Arabic (ar)   & 157.28  & 91.74  & 58.3\% & 136{,}040   & 1{,}651 \\
\hline
\end{tabular}
\caption{Cross-language comparison of Mozilla Common Voice v24.0 dataset-card statistics. Validation rate is computed as validated hours divided by total recorded hours. Source: Mozilla Common Voice v24.0 dataset cards (retrieved 14 Feb 2026) ~\cite{ref11}.}
\label{tab:cv_cross_language_v24}
\end{table}

As shown in Table~\ref{tab:cv_cross_language_v24}, Pashto has transitioned from severe data scarcity to large-scale availability within the Common Voice ecosystem. However, total duration alone does not characterize dataset structure. The following sections examine how participation patterns, validation throughput, and metadata availability shape the internal composition of the corpus.

\section{Dataset Overview and Methodology}

This section describes the data source, structural organization, and analytical framework used in this study. All statistics are computed directly from the official release files of the Pashto subset of Mozilla Common Voice (v24.0).

\subsection{Data Source}

The Pashto dataset is part of the Mozilla Common Voice initiative, a multilingual, community-driven speech collection project released under the CC0 1.0 Universal license~\cite{ref3}. Contributors record prompted sentences, and other participants validate the recordings through peer review.

We analyze release versions v14.0, v20.0, and v24.0 for longitudinal comparison. Unless otherwise specified, descriptive statistics refer to v24.0.

The following files were used for analysis:

\begin{itemize}
    \item \texttt{validated.tsv}
    \item \texttt{invalidated.tsv}
    \item \texttt{other.tsv}
    \item \texttt{reported.tsv}
    \item \texttt{train.tsv}, \texttt{dev.tsv}, \texttt{test.tsv}
\end{itemize}

All clip-level analyses are conducted on the \texttt{validated.tsv} subset unless explicitly stated otherwise.

\subsection{Dataset Composition (v24.0)}

As of version 24.0 (December 2025), the Pashto dataset contains:

\begin{itemize}
    \item 2,407,799 total recorded clips
    \item 2,768.7 total recorded hours
    \item 975.89 validated hours
    \item 6,654 unique speakers (based on \texttt{client\_id})
    \item 59,369 unique sentences
\end{itemize}

Duration statistics are computed by summing clip durations (milliseconds) and converting to hours. Speaker counts are derived from unique \texttt{client\_id} entries in \texttt{validated.tsv}.

\subsection{Validation and Subset Structure}

Each clip undergoes community validation through up-vote and down-vote mechanisms. Clips receiving sufficient positive validation are included in the \textit{validated} subset. Clips may also appear in the following partitions:

\begin{itemize}
    \item \textbf{Validated} – clips approved through peer review
    \item \textbf{Invalidated} – clips rejected during validation
    \item \textbf{Other} – clips pending sufficient validation
    \item \textbf{Reported} – clips flagged for potential issues
\end{itemize}

Official modeling splits (Train, Dev, Test) are derived from the validated subset. In v24.0, 19.91\% of validated clips are included in these benchmark splits, reflecting predefined evaluation constraints rather than exhaustive allocation of all validated data.

\subsection{Units of Analysis}

The primary unit of analysis is the audio clip. Speaker-level statistics are computed by aggregating validated clips by \texttt{client\_id}. Sentence-level statistics are computed using the \texttt{sentence} field in \texttt{validated.tsv}.

Unless otherwise stated:

\begin{itemize}
    \item Participation inequality is computed over validated clips only.
    \item Sentence concentration metrics are computed over validated clips.
    \item Demographic analyses are restricted to entries with non-empty metadata fields.
\end{itemize}

\subsection{Metadata Handling}

Age, gender, and accent fields are optional and self-reported. Missing values are treated as an explicit \textit{Undefined} category and are not imputed. Because disclosure is voluntary, demographic distributions reflect both participation patterns and reporting behavior.

Domain metadata is reported descriptively due to incomplete and inconsistent labeling across releases.

\subsection{Reproducibility}

All analyses were conducted using Python (pandas, numpy, matplotlib).
The raw dataset files analyzed in this study are publicly available
through the Mozilla Data Collective:
\url{https://datacollective.mozillafoundation.org/datasets/cmj8u3pnb00llnxxbfvxo3b14}
(accessed 14 Feb 2026).

All reported statistics can be reproduced by parsing the release
metadata files (validated.tsv, invalidated.tsv, other.tsv,
reported.tsv, train.tsv, dev.tsv, test.tsv) included in version 24.0.
Analysis scripts will be publicly archived in a version-controlled
repository accompanying this work.

\section{Dataset Evolution and Quantitative Growth Analysis}

This section examines the longitudinal evolution of the Pashto Common Voice dataset, focusing on scale expansion, validation dynamics, and participation structure. These analyses contextualize both the strengths and structural constraints of the December 2025 release (v24.0).

\subsection{Longitudinal Growth Across Releases}

The Pashto corpus has undergone rapid expansion across three key milestones: version 14.0 (June 2023), version 20.0 (December 2024), and version 24.0 (December 2025).

\begin{itemize}
    \item \textbf{Initial Scarcity (v14.0, June 2023):} 1.49 total hours, of which 1.27 hours were validated.
    \item \textbf{Intermediate Expansion (v20.0, December 2024):} 115.13 total hours, including 64.36 validated hours.
    \item \textbf{Large-Scale Availability (v24.0, December 2025):} 2,768.7 total recorded hours, including 975.89 validated hours.
\end{itemize}

\begin{figure}[ht]
    \centering
    \includegraphics[width=0.75\linewidth]{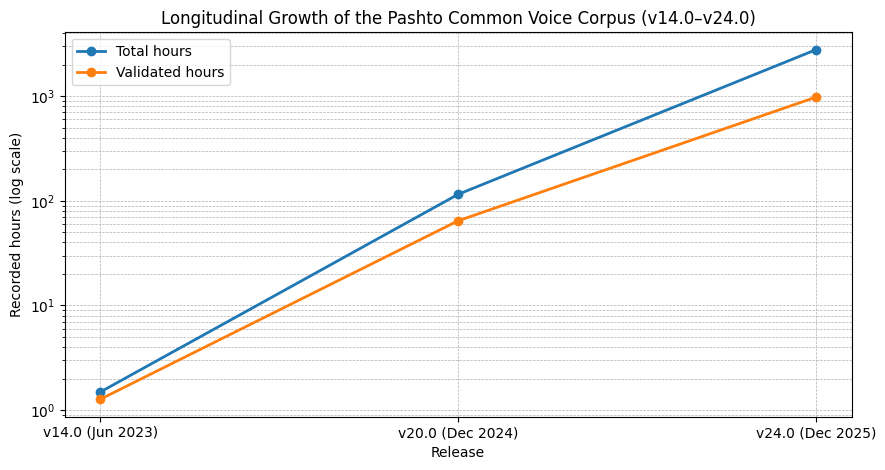}
    \caption{Growth of the Pashto Common Voice corpus across major releases, showing total and validated hours from June 2023 (v14.0) to December 2025 (v24.0) on a logarithmic scale.}
    \label{fig:growth}
\end{figure}

Figure~\ref{fig:growth} illustrates this exponential trajectory on a logarithmic scale. The transition from fewer than two hours in 2023 to nearly 2,800 hours in 2025 represents a structural transformation from extreme data scarcity to large-scale availability within less than three years.

Importantly, validated hours increased from 1.27 in v14.0 to 975.89 in v24.0, demonstrating not only recording expansion but also substantial growth in training-ready data. Nevertheless, rapid scaling raises questions regarding representational balance, validation throughput, and participation concentration, issues examined in the following subsections.

\subsection{Validation Dynamics and Practical Usability}

According to the v24.0 dataset card ~\cite{ref11}, the Pashto Common Voice corpus contains 2,407,799 total clips corresponding to 2,768.7 hours of recorded speech. Of this volume, 975.89 hours are validated, representing 35.2\% of total recorded duration.

At the clip level, 687,158 clips are validated, 41,244 are invalidated, and 1,571,559 remain in the \textit{other} category (i.e., awaiting sufficient peer validation). Although the validated subset now provides a substantial volume suitable for supervised ASR training, most recorded clips are not yet included in the training-ready pool.

This distribution reflects a common pattern in community-driven collection systems: recording activity scales more rapidly than validation. Validation requires additional participant effort and repeated review, which can constrain throughput relative to raw recording growth.

Importantly, the size of the \textit{other} category does not imply low recording quality. Instead, it indicates that a large portion of contributed audio remains unverified and therefore unavailable for supervised training. Increasing validation throughput would directly expand the usable portion of the corpus without requiring additional recording.

\subsection{Contribution Inequality and Participation Structure}
\label{sec:participation_structure}

The v24.0 release contains contributions from 6,654 unique speakers
(as identified by distinct \texttt{client\_id} values in the validated subset). 
However, participation volume is highly unevenly distributed.

\begin{figure}[ht]
    \centering
    \includegraphics[width=0.8\linewidth]{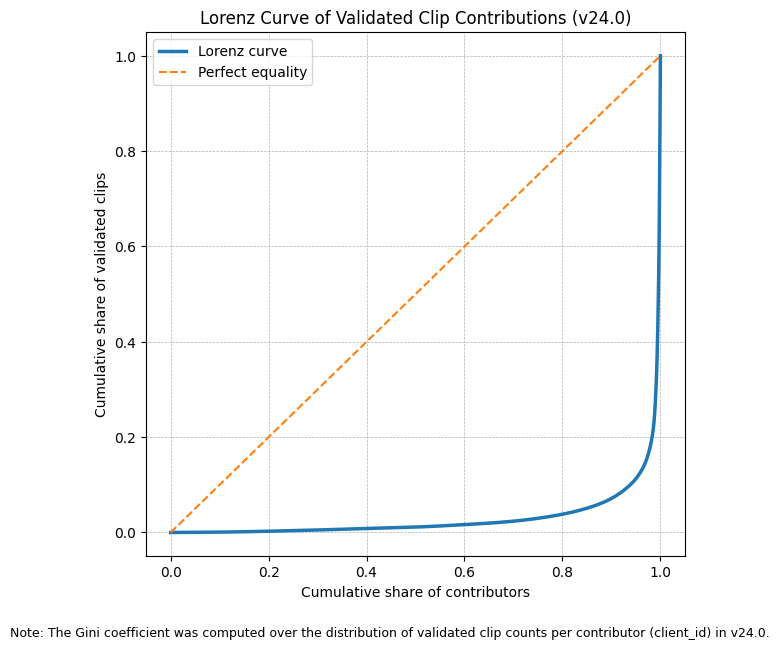}
    \caption{Lorenz curve of validated clip contributions across contributors in the Pashto Common Voice v24.0 release. The corresponding Gini coefficient (0.941) indicates a highly unequal, long-tail contribution structure. The Gini coefficient was computed over the distribution of validated clip counts per contributor (client\_id ) in version 24.0.}
    \label{fig:lorenz_curve}
\end{figure}

To quantify contribution concentration, we use the Lorenz curve and the Gini coefficient, 
a standard measure of distributional inequality~\cite{ref13}. 
Figure~\ref{fig:lorenz_curve} shows the Lorenz curve for validated clip contributions. 
The curve departs strongly from the line of perfect equality, yielding a Gini coefficient 
of $G = 0.941$, indicating extreme concentration of validated speech among a small subset 
of contributors.

The Gini coefficient is computed from the Lorenz curve as:
\[
G = 1 - 2 \int_{0}^{1} L(p)\, dp
\]
where $L(p)$ denotes the cumulative share of validated clips contributed by the bottom 
$p$ proportion of speakers.

Such long-tail participation patterns are common in crowdsourced platforms, where a small 
number of highly active contributors account for a disproportionate share of output. 
While these contributors enable rapid scaling (Figure~\ref{fig:growth}), reliance on a 
limited subset of speakers can reduce acoustic diversity and skew representational coverage.

As a result, increasing total hours alone is insufficient; targeted recruitment and outreach 
may be required to broaden speaker diversity and reduce dependence on a small number of 
high-activity contributors.

\section{Descriptive Statistics and Dataset Composition}

This section presents a quantitative snapshot of the December 2025 release (v24.0), establishing a statistical baseline for subsequent demographic, thematic, and governance analyses. The objective is to characterize not only aggregate scale but also the structural properties that influence downstream ASR performance.

\subsection{Corpus Scale and Validation Structure}

The corpus comprises 2,407,799 speech clips totaling 2,768.7 hours of recorded audio. Of this volume, 975.89 hours have passed peer validation and are therefore considered training-ready.

The dataset is organized into four primary validation states:

\begin{itemize}
    \item \textbf{Validated} – clips approved through community peer review
    \item \textbf{Invalidated} – clips rejected during validation
    \item \textbf{Other} – clips awaiting sufficient validation
    \item \textbf{Reported} – clips flagged for potential quality concerns
\end{itemize}

While validated audio now constitutes a substantial resource suitable for modern ASR training, a large proportion of contributed clips remains in the \textit{other} category, reflecting the structural validation backlog characteristic of community-driven pipelines.

\subsection{Official Modeling Splits}

From the validated subset, Common Voice provides standardized Train, Dev, and Test splits to support reproducible evaluation. Importantly, only 19.91\% of validated clips are included in these official splits.

This design reflects a benchmarking-oriented philosophy: researchers are encouraged to construct custom training sets from the full validated pool while reserving official splits for stable cross-study comparison. The relatively modest size of the official splits should therefore be interpreted not as a limitation, but as a deliberate evaluation protocol.

\subsection{Text Corpus Structure}

The Pashto text corpus contains 266,477 unique sentences. These are distributed as follows:

\begin{itemize}
    \item 246,535 validated sentences
    \item 15,712 invalidated sentences
    \item 4,230 reported sentences
\end{itemize}

Because each sentence may be recorded by multiple speakers, the dataset exhibits a many-to-one relationship between clips and prompts. This structure promotes acoustic variability while potentially limiting textual diversity, a trade-off further analyzed in the following section.

\subsection{Speaker Participation and Duration Characteristics}

Version 24.0 includes contributions from 6,654 unique speakers. Despite this broad participation base, contribution volume is unevenly distributed (see Section~\ref{sec:participation_structure}).

The average validated clip duration is approximately 4.1 seconds, consistent with the platform’s short-utterance design optimized for ease of validation and community participation.

\begin{figure}[ht]
    \centering
    \includegraphics[width=0.75\linewidth]{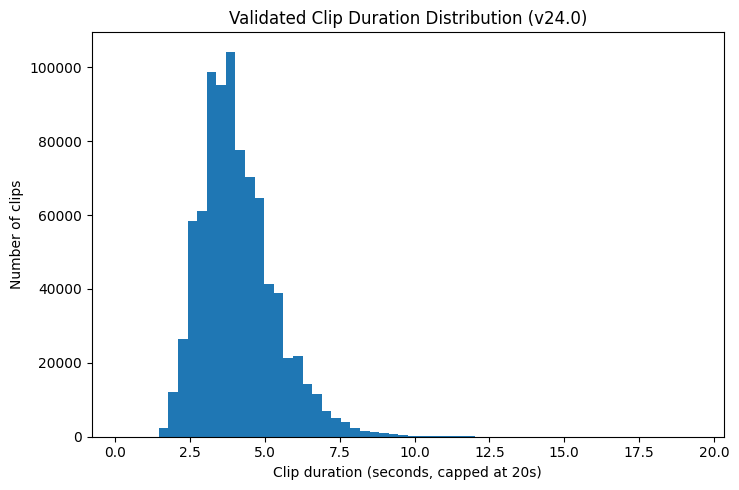}
    \caption{Distribution of validated clip durations in the Pashto Common Voice v24.0 release. Durations are capped at 20 seconds for readability, reflecting the short-utterance design of the Common Voice platform.}
    \label{fig:clip_duration}
\end{figure}

\subsection*{Interpretation}

Taken together, these statistics indicate that v24.0 provides nearly 1,000 hours of validated speech, sufficient to support modern neural ASR training pipelines. At the same time, the relatively modest proportion of validated data relative to total contributions underscores the importance of sustained validation engagement. Furthermore, the relationship between clip count, sentence repetition, and participation inequality suggests that scale alone does not guarantee proportional growth in textual or demographic diversity.

\section{Demographic Composition Analysis}

Demographic composition plays a central role in both generalization and fairness in speech recognition systems. Models trained on skewed datasets may achieve high aggregate performance while underperforming for underrepresented speaker groups. This section analyzes clip-level age and gender metadata in v24.0. Because demographic disclosure is optional and self-reported, the distributions reflect both participation patterns and willingness to provide personal information.

\subsection{Age Distribution}

\begin{table}[ht]
\centering
\caption{Age distribution of clips (v24.0)}
\begin{tabular}{lrr}
\hline
Age Category & Clips & Percentage \\
\hline
Twenties & 1,433,630 & 59.53\% \\
Thirties & 472,795 & 19.63\% \\
Teens & 191,562 & 7.96\% \\
Forties & 90,848 & 3.77\% \\
Fifties & 83,885 & 3.48\% \\
Sixties & 2,123 & 0.09\% \\
Seventies & 89 & $<$0.01\% \\
Undefined & 132,867 & 5.52\% \\
\hline
\end{tabular}
\end{table}

The distribution reveals a strong concentration among young adult speakers. Contributors in their twenties and thirties account for approximately 79.16\% of all clips, while speakers aged sixty and above represent less than 0.1\% of the dataset.

This imbalance has important implications for ASR robustness. Age correlates with acoustic and prosodic variation, including differences in pitch stability, articulation rate, and vocal quality. Limited representation of older speakers may therefore constrain model generalization to elderly populations.

From a socio-technical perspective, this skew likely reflects patterns of digital participation: crowdsourced platforms disproportionately attract younger, digitally connected contributors. Without targeted outreach, such demographic concentration may become structurally embedded.

\subsection{Gender Distribution}

\begin{table}[ht]
\centering
\caption{Gender distribution of clips (v24.0)}
\begin{tabular}{lrr}
\hline
Gender Category & Clips & Percentage \\
\hline
Female Feminine & 1,390,861 & 57.76\% \\
Male Masculine & 6,244 & 0.26\% \\
Do Not Wish To Say & 212 & $<$0.01\% \\
Undefined & 1,010,482 & 41.97\% \\
\hline
\end{tabular}
\end{table}

The dataset contains a substantial proportion of clips labeled as Undefined (41.97\%), reflecting optional demographic disclosure. Among labeled entries, female-labeled contributions constitute the majority, while male-labeled clips represent a very small share.

These figures require cautious interpretation. Optional metadata design protects privacy and reduces participation barriers but introduces structural ambiguity in demographic assessment. To better understand the Undefined category, we conducted an exploratory perceptual inspection of a randomly sampled subset ($N=300$) drawn from clips without gender labels. Blinded listening suggested that many recordings exhibited acoustically male vocal characteristics. This observation is illustrative and not intended as demographic inference, but indicates that metadata incompleteness may not fully reflect acoustic participation patterns.

Prior research has demonstrated that ASR systems frequently exhibit performance disparities across gendered speaker groups when training data is imbalanced~\cite{ref5,ref6}. Even incomplete demographic metadata therefore provides important signals for fairness risk assessment.

\section{Domain Metadata and Sentence-Level Concentration}
This section examines the thematic and textual coverage of the Pashto Common Voice corpus. In read-speech datasets, the diversity of recording prompts strongly influences vocabulary coverage and linguistic variability. We therefore analyze the available domain metadata and the distribution of recordings across unique sentences.
\subsection{Domain Metadata Structure}

The Pashto Common Voice release includes sentence-level domain labels (e.g., \textit{General}, \textit{News}, \textit{Education}, \textit{Technology}). However, a substantial proportion of entries are marked as \textit{Undefined}, and the assignment criteria for these categories are not formally documented.

Because domain labels are sparsely populated and inconsistently specified across releases, we do not perform a quantitative domain distribution analysis. The high prevalence of undefined labels limits reliable assessment of topical balance. As a result, domain metadata is treated descriptively rather than as a stable basis for statistical evaluation.

\subsection{Sentence-Level Concentration}

\begin{figure}[t]
    \centering
    \includegraphics[width=0.8\linewidth]{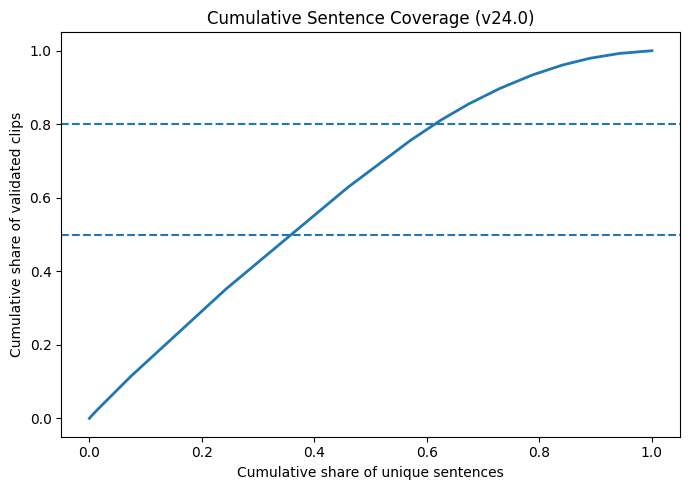}
    \caption{Cumulative coverage of validated clips by unique sentences in the Pashto Common Voice v24.0 release, illustrating the concentration of recordings across a relatively small subset of prompts.
}
    \label{fig:sentence_coverage}
\end{figure}

Textual diversity can instead be assessed through repetition patterns within the validated subset. Figure~\ref{fig:sentence_coverage} presents the cumulative share of validated clips as a function of the cumulative share of unique sentences.

The distribution indicates moderate prompt reuse. Specifically, 35.88\% of unique sentences account for 50\% of validated clips, while 61.37\% account for 80\%. Although some prompts are recorded more frequently than others, validated clips are distributed across a broad portion of the available sentence inventory.

This pattern contrasts with the extreme speaker-level inequality documented in Section~\ref{sec:participation_structure} (Gini = 0.941), suggesting that concentration in the corpus arises primarily from uneven contributor activity rather than dominance of a small set of prompts.

\section{Discussion and Conclusion}

The Pashto Common Voice corpus has expanded from fewer than two recorded hours in 2023 to 2,768.7 total hours in version 24.0, including 975.89 validated hours contributed by 6,654 speakers. This growth substantially improves the availability of openly licensed Pashto speech data for ASR development. However, scale alone does not determine representational balance or structural robustness.

Our analysis identifies three principal structural characteristics.

First, participation is highly concentrated. A small subset of contributors accounts for a disproportionate share of validated clips (Gini = 0.941; Section~\ref{sec:participation_structure}). Such inequality suggests that corpus growth is driven primarily by repeated contributions rather than broad speaker expansion. While this does not reduce total duration, it may limit acoustic diversity relative to raw scale.

Second, demographic metadata is incomplete. A substantial proportion of clips lack gender labels, and older speakers are sparsely represented. Because demographic disclosure is optional, metadata distributions reflect both participation patterns and reporting behavior. This limits fine-grained subgroup analysis and constrains direct fairness auditing based solely on metadata fields.

Third, validation throughput lags behind recording volume. Only 35.2\% of total recorded hours are included in the validated subset. The remaining audio constitutes latent training capacity that is currently excluded from supervised ASR training pipelines.

Taken together, these findings indicate that the Pashto corpus has transitioned from scarcity to scale, but structural composition remains uneven. Dataset maturity may benefit from broader contributor diversity, improved metadata completeness, and increased validation throughput.

\subsection{Implications}

For ASR development, the present structure supports large-scale pretraining and benchmarking, particularly for read-speech scenarios. However, participation inequality and metadata gaps may influence demographic generalization and robustness across age groups.

Future improvements could prioritize:

\begin{itemize}
    \item Expanding validation capacity to reduce backlog.
    \item Encouraging participation from underrepresented age groups.
    \item Improving optional metadata reporting while preserving contributor privacy.
\end{itemize}

Such adjustments would enhance representational balance without requiring fundamental changes to the crowdsourcing model.

\section{Data Availability}

The Mozilla Common Voice Pashto dataset (v24.0) is publicly available under a CC0 1.0 public-domain dedication at \url{https://commonvoice.mozilla.org}. All analyses in this study rely exclusively on publicly released metadata and audio files.

\section{Ethics Statement}

This study uses publicly available data released under a CC0 license~\cite{ref3,ref12}. All contributors consent to public distribution of their recordings through the Common Voice platform. Our analysis relies solely on anonymized metadata and does not attempt to identify individual speakers.

We note that demographic imbalances and incomplete metadata may influence downstream ASR performance. By documenting these structural characteristics, this work aims to support transparent evaluation and responsible model development.

\end{document}